\documentclass[11pt,letterpaper]{article}
\usepackage{cogsys}
\usepackage[T1]{fontenc}
\usepackage{times}
\usepackage[pdftex]{graphicx} 
\usepackage{url}  

\usepackage[]{nowidow}
\usepackage{array}
\usepackage{rotating}
\usepackage{natbib}
\usepackage{booktabs}
\usepackage{multirow,bm}
\usepackage{adjustbox}
\usepackage{makecell}
\usepackage{tablefootnote}
\usepackage[para,online,flushleft]{threeparttable}

\cogsysheading{X}{20XX}{1-6}{X/20XX}{X/20XX}

\ShortHeadings{Reflecting After Learning for Understanding}
              {L.\ Martie, M.\ Alam, G. Zhang, and R.\ Anderson}

\begin{document} 

\title{Reflecting After Learning for Understanding}
 
\author{Lee Martie}{lee.martie@ibm.com}
\author{Mohammad Arif Ul Alam}{arif.alam@ibm.com}
\author{Gaoyuan Zhang}{gaoyuan.zhang@ibm.com}
\address{MIT-IBM Watson AI Lab, IBM Research, 75 Binney Street, Cambridge, MA 02142 USA}
\author{Ryan R.\ Anderson}{rranders@us.ibm.com}
\address{IBM Cloud and Cognitive Software, 505 Howard St, San Francisco, CA 94105 USA}
\vskip 0.2in
 
\begin{abstract}Today, image classification is a common way for systems to process visual content. Although neural network approaches to classification have seen great progress in reducing error rates, it is not clear what this means for a cognitive system that needs to make sense of the multiple and competing predictions from its own classifiers. As a step to address this, we present a novel framework that uses meta-reasoning and meta-operations to unify predictions into abstractions, properties, or relationships. Using the framework on images from ImageNet, we demonstrate systems that unify 41\% to 46\% of predictions in general and unify 67\% to 75\% of predictions when the systems can explain their conceptual differences. We also demonstrate a system in ``the wild'' by feeding live video images through it and show it unifying 51\% of predictions in general and 69\% of predictions when their differences can be explained conceptually by the system. In a survey given to 24 participants, we found that 87\% of the unified predictions describe their corresponding images. 
\end{abstract}

\section{Introduction} 
 
Learning in artificial intelligence is most often framed as training one or more flavors of deep neural networks \citep{goodfellow_deep_2016}. Advances in image classifiers, for example, rely on convolutional neural networks or generative adversarial networks \citep{goodfellow_generative_2014}. Learning identity functions through autoencoders often rely on training a multilayer perceptron or more advanced techniques \citep{goodfellow_autoencoders_2016}. While reinforcement learning can use tables \citep{sutton_reinforcement_2018}, scalable solutions often use deep neural networks, as in \citeauthor{mnih_playing_2013}'s \citeyearpar{mnih_playing_2013} work on playing Atari video games.

In brief, a deep neural network for image classification is often created by tuning weights in a neural network structure to optimize a loss function, such that one label from a given set is predicted as most likely for an input. One consequence is that the understanding of images has been narrowed to predicting a most likely label, given some input from a data distribution. Some well-known examples of this are ImageNet classifiers, such as AlexNet \citep{krizhevsky_imagenet_2012}, ResNet \citep{he_deep_2016}, and SqueezeNet \citep{iandola_squeezenet:_2016}, which have been trained on millions of images over millions of parameters to predict the ``correct'' label from a thousand alternatives. Indeed, ImageNet held a competition where classifiers were compared with each other according to their error rate of predicting the correct label (top--1 error rate) and the error rate of predicting the correct label in the top five predictions (top--5 error rate).

\begin{figure}[t]
\vskip 0.05in
\begin{center}
\includegraphics[width=3.5in]{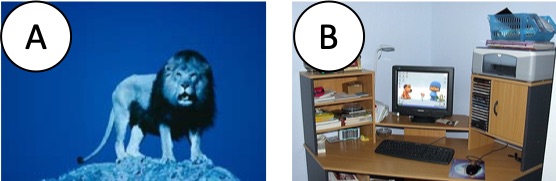}
\caption{Image A and B from ImageNet, where A is classified as ``arctic fox'' by AlexNet and ``ox'' by ResNet, and B is classified as ``desk'' by AlexNet and ``desktop computer'' by ResNet.} 
\label{lion-desk}
\end{center}
\vskip -0.2in
\end{figure} 

While ImageNet classifiers are reducing their error rates, it is not clear what error rate reduction says about making sense of visual input. In particular, it remains unclear how a cognitive system might know what it is looking at, especially after deployment when it cannot check ``correctness'' against a test set.  Further, while classifiers are traditionally pitted against one another in order to find the best \citep{kaggle_inc_competitions_2019}, we observe they often can output different ``correct'' labels after learning, regardless of what the testing data suggests. For example, we took images A and B in Figure~\ref{lion-desk} from ImageNet \citep{stanford_vision_lab_imagenet_2019} and asked AlexNet and ResNet to classify it. For Image A, AlexNet classifies this image as ``arctic fox,'' and ResNet classifies this image as ``ox''. Strictly speaking, both classifiers are incorrect, and neither would know in production. However, in some sense they are also correct (i.e., both an arctic fox and ox are mammals and a mammal does appear in the image). For Image B, AlexNet classifies this image as ``desk,'' and ResNet classifies this image as ``desktop computer,'' but the correct label, according to the ImageNet challenge, is ``desk'' \citep{stanford_vision_lab_imagenet_2012}. Again, both classifiers are correct in some sense (i.e., both a desk and desktop computer occur in the image). As such, labels are more like views of what is being seen and point to some encoded knowledge rather than a ``correct'' description. In this paper, we adopt the word \textit{view} to mean the encoded knowledge in a classifier that the predicted label loosely describes.

Recent research acknowledges one label for a picture is limiting and multi-label classifiers, such as YOLO \citep{redmon_you_2016}, provide more labels per image. However, a critical problem remains. How can complementary or competing views, after they have already been learned, be combined into some kind of higher-level knowledge that can be understood and reasoned over? Such concept combinations appears to be a key component in how humans think (e.g., creating abstractions and relationships) \citep{bayne_what_2003, shivhare_cognitive_2016} and is critical for using visual input to establish correct conditions for planning and verifying achievement of goals. As such, this is a key question for understanding and building cognitive systems that include frameworks for sense-making.

In this paper, we investigate the research question \textit{How can a cognitive system's different views be unified into higher-level knowledge after learning?} We address this question and the value of such a system with four main contributions:
\vskip 0.05in

\cbullet 
We present a novel approach and framework for creating cognitive systems that can reconcile views through a process of convergence, where multiple views are collected and unified into abstractions, relationships, or properties by leveraging reflection. We call our overall approach \textit{symbolic mirroring} and refer to the paradigm as the \textit{symbolic mirroring framework} (SMF).

\cbullet 
Through a detailed planning example, we demonstrate the value of our approach by showing how a system can unify its views (correct or not) to find a level of abstraction that identifies conditions when executing a plan. 

\cbullet 
We evaluate how well our approach can unify views in an exploratory laboratory study, where we evaluate three different systems, in our framework, over 950 images from ImageNet. We find that the systems can unify a substantial number of views when their differences can be explained by differences in their abstractions. In particular, we find the programs were able to unify views for 41\% to 46\% of the images and, among the views the programs could explain, they could unify 67\% to 75\% (all $p \leq .001$). In an exploratory field study, we find a system in our framework can unify 51\% of the views created from live video images and, among the views that it could explain, the system could unify 69\% ($p = .041$). 

\cbullet 
In a survey given to 24 participants, we find they agreed that 87\% ($p \leq .001$) of the unified views made by a system in our framework describe their corresponding images. 

\vskip 0.05in
\noindent
The construction of the cognitive systems we evaluate utilize several novel concepts introduced in our symbolic mirroring framework (SMF). In particular, our framework introduces \textit{meta-points}, which supports reflection on executed and unexecuted portions of a cognitive system by passing parts of the system to higher-level operations called \textit{meta-operations}. The meta-operation \textit{explain} provides an explanation when multiple views are not the same and the meta-operation \textit{converge} unifies these views into higher-level concepts through an algorithm on domain knowledge specified in the OWL description logic \citep{horrocks_owl:_2005} and using the OWL reasoner, Pellet \citep{sirin_pellet:_2007}. In this way, meta-points and meta-operations act to bridge views from trained neural networks to higher-level knowledge in symbolic approaches.

The rest of the paper provides the details of our research. Section 2 introduces our approach and framework. Section 3 demonstrates why symbolic mirroring is useful for planning. Section 4 describes the experiment design for evaluating our approach, presents the results, and discusses threats to validity. Section 5 discusses the results, Section 6 contrasts our work with related work, and Section 7 concludes with a view on future work.

\begin{figure}[t]
\vskip 0.05in
\begin{center}
\includegraphics[width=5.5in]{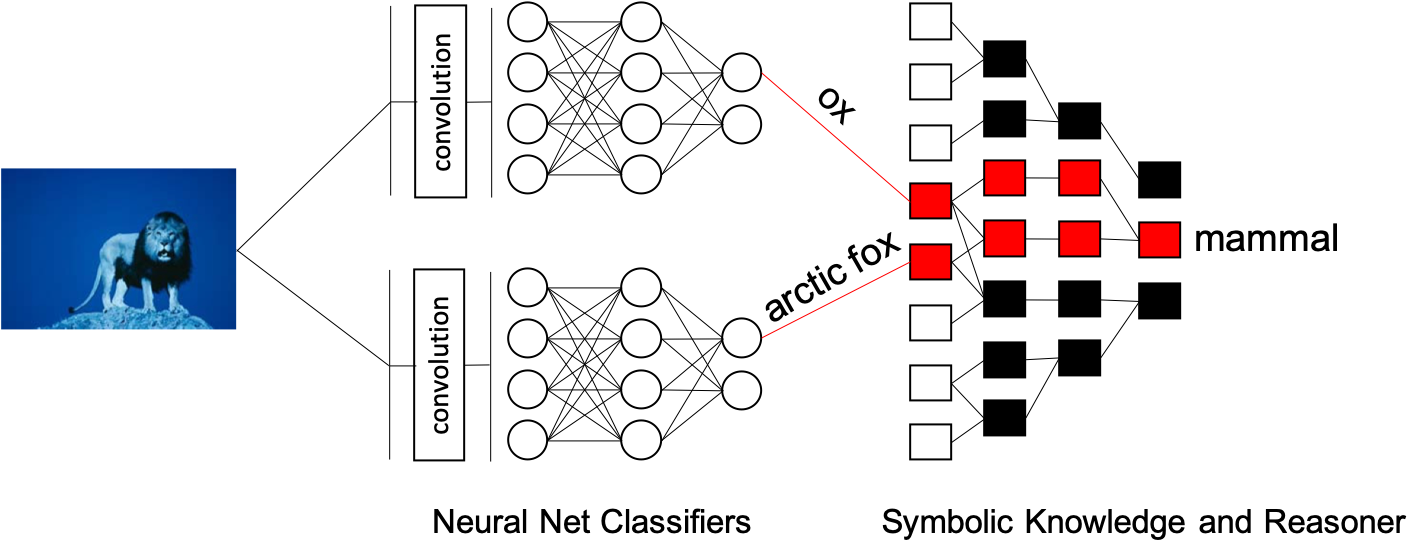}
\caption{Example of our symbolic mirroring architecture, where knowledge in classifiers converges into a higher level of abstraction.} 
\label{symbolic-mirror}
\end{center}
\vskip 0.1in
\end{figure} 

\section{Symbolic Mirroring Approach and Framework}

We designed our framework for building cognitive systems that can unify different views among their classifiers into higher-level knowledge through a process of convergence called \textit{symbolic mirroring}. The conjecture behind our approach is that convolutional neural networks contain hidden knowledge describing input images, even when their predictions are incorrect. While knowledge in neural networks suffers from an opacity problem, the symbolic mirroring technique attempts to make this knowledge explicit for an input. The approach involves creating a knowledge base that expresses abstractions, properties, and relationships in the domain that the classifiers' labels reside in order to ``mirror'' some of the knowledge inside the classifiers -- hence the name symbolic mirroring. The top predictions from two classifiers are then mapped into the knowledge base to find common abstractions, properties, and relationships. Our approach looks for common higher-level knowledge in order to converge on concepts that the predictions of the classifiers suggest exists in the image. Figure~\ref{symbolic-mirror} presents an example of a symbolic mirroring architecture that our framework supports. In this architecture, the program attempts to unify what each classifier understands about the image by mapping their top labels (even if both are strictly wrong) onto instances (white boxes) in a knowledge base, where a reasoner finds the closest ancestor (abstraction) of both instances. In this example, the system decides it is looking at a mammal rather than an ox or an arctic fox.

The design of programs in our framework follows a composable graph design pattern, similar to the composite and command pattern \citep{gamma_design_1995}, but where a node in the graph has a function and annotations map the node into meta-knowledge (so it can be reasoned over at runtime). A directed edge from one node to the next specifies the execution order. When a node is executed, its function is run and the return value of the function is assigned to the node. The value of one node (or any assigned to it in the past) can be retrieved by another node's function downstream as an argument. The final specified graph is executed by the framework's runtime. 

\begin{figure}[t]
\vskip 0.05in
\begin{center}
\includegraphics[width=5.2in]{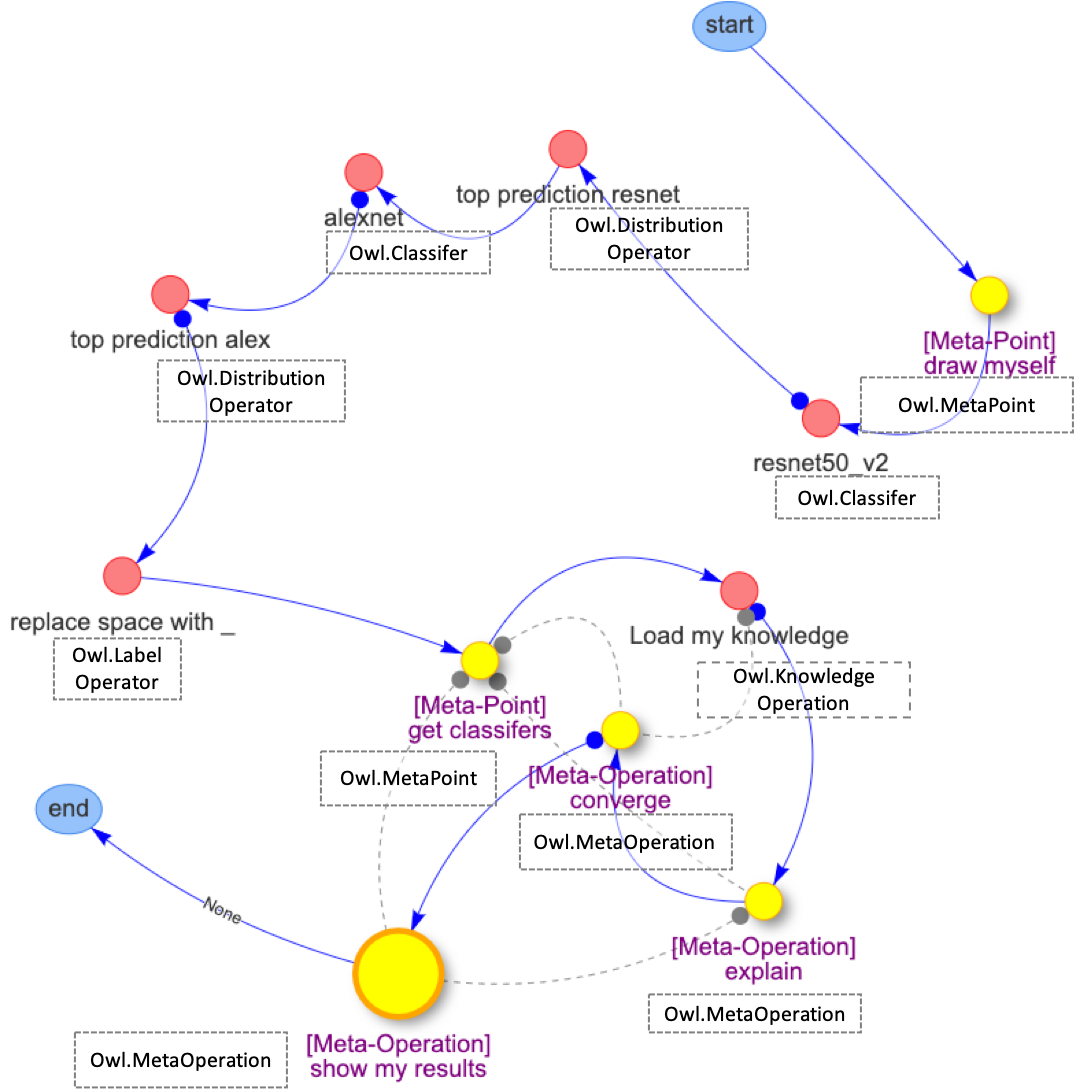}
\caption{Auto-created graph of executing SMF program, where views of ResNet and AlexNet converge into higher-level knowledge through reflection.} 
\label{meninx-graph}
\end{center}
\vskip -0.2in
\end{figure} 

Two key node types in the framework are utilized in our programs to unify knowledge. In particular, these are meta-points and meta-operations. Meta-points are nodes that take the entire graph as input in order to perform some function (e.g., finding nodes of a particular type). Meta-points support easily extending programs in a modular way, because they simply can be added to any program and perform higher-level reasoning about it without the programmer needing to parameterize the node for it to work with the rest of the program. Meta-operations perform some operations on parts of the graph, but do not take the graph as input. For example, a meta-operation might make changes to the program's knowledge and/or graph structure but relies on arguments from meta-points or other nodes. We next walk through the features of the framework illustrated with an example.

\subsection{Features of the Symbolic Mirroring Framework}

We designed an API that supports building SMF programs as first-class entities in Python so the program itself can be reasoned over by the other meta-reasoning components in the program. Consider the graph visualization of an executed program shown Figure~\ref{meninx-graph}, where we annotated the nodes with their OWL class in the meta-knowledge (these appear in white boxes). The SMF runtime executes the program in Figure~\ref{meninx-graph} starting at the start node, which simply transitions to the next node to execute (indicated by an arrow pointing to the next node). The next node to execute is the node \textit{draw\_self} (a meta-point). \textit{Draw\_self} takes the entire program graph as input (as do all meta-points) and then calls a function to draw the entire program (including the node \textit{draw\_self}) to a Web application (Figure~\ref{meninx-graph} is the result). The next node to execute is the \textit{resnet50\_v2 node}, which feeds the image B shown in Figure~\ref{lion-desk} to a trained ResNet classifier via a REST request. The resulting distribution is assigned to the node. The \textit{top prediction resnet} node retrieves the distribution of predictions from the \textit{resnet50\_v2} node (indicated by the edge ending in a circle) and executes a function to get the top prediction (in this case ``desktop computer''). The next node to be executed is the \textit{alexnet} node which also feeds image B shown in Figure~\ref{lion-desk} to a trained AlexNet classifier via a REST request. The value of the \textit{alexnet} node is another distribution of predictions which the \textit{top prediction alex} node retrieves and, in turn, produces the top prediction (in this case ``desk''). This prediction is retrieved and syntactically modified by the \textit{replace space with \_} node.

\begin{figure}[t]
\vskip 0.05in
\begin{center}
\includegraphics[width=6in]{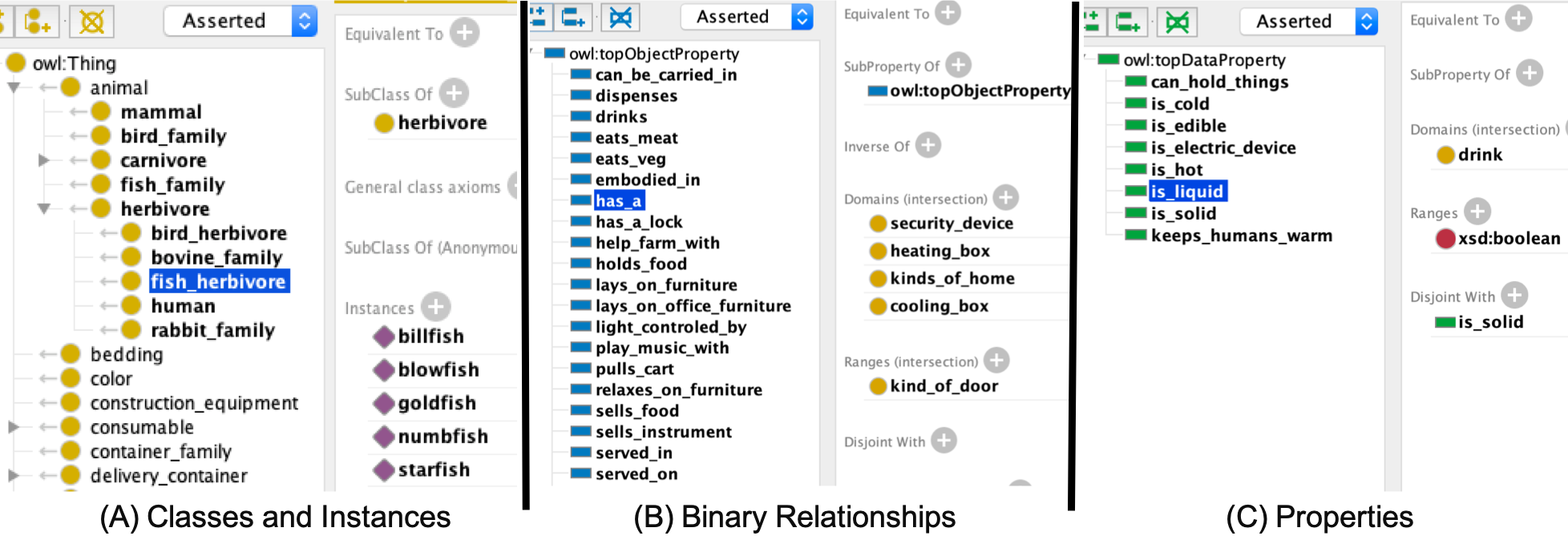}
\caption{Sample of ImageNet domain knowledge in OWL as displayed in Prot\'eg\'e.} 
\label{knowledge-protege}
\end{center}
\vskip -0.2in
\end{figure}

The next node to execute is another meta-point named \textit{get classifiers} and, as such, the entire graph gets fed into this node. The \textit{get classifiers} meta-point finds the values of executed classifiers by looking at the preceding executed graph and locating nodes that are annotated with the \textit{OWL.Classifier} class. Once the classifiers are identified, the relevant parts of the executed program are retrieved downstream by the \textit{converge}, \textit{explain}, and \textit{show my results} nodes (this is indicated by edges ending with a circle, where dashed edges indicate only a dependency exists). But the next node to be executed is the \textit{Load my knowledge} node, which loads domain knowledge for ImageNet classes, which is later used to reason about the classifiers' views. Since ImageNet labels describe common-sense-kind of concepts (e.g., computer, dog, and jacket), one author was able to create domain knowledge specifying the abstractions, properties and relationships of the ImageNet domain in OWL using Prot\'eg\'e \citep{stanford_center_for_biomedical_informatics_research_protege_2019}.  Figure~\ref{knowledge-protege} shows a portion of this knowledge in the Prot\'eg\'e interface. In total, we created 139 classes (abstractions), 25 relationships, and nine properties. Each individual in the ontology can be assigned to a class or have inferred classes and can have one or more relationships with another instance (called \textit{object properties} in OWL) and have one or more properties (called \textit{data properties} in OWL).

Once the ontology is loaded, the next node to be executed is the explain node. This meta-operation explains why different views do not match in terms of their OWL classes. The way it does this is to first map the views to individuals in the ontology by mapping their labels to individuals. That is, it treats views as instances of concepts that it knows about. If no concept is found, no explanation can be made. If the mapping is made, the explain node looks to see if the instances are the same or not. If not, it returns an explanation about how they differ by class. In our example, the explain node outputs ``desk is a kind of furniture and desktop computer is a kind of computing device.'' Next, the \textit{converge} node executes by taking in the parts of the program from the meta-point \textit{get classifiers} and the domain knowledge from the \textit{Load my knowledge} node. Here, the \textit{converge} node runs the converge algorithm that attempts to unify the views from the classifiers into higher-level knowledge. Because the converge algorithm reasons about the system's inferences about what it is seeing (i.e., the classifiers' views), the converge nodes is a type of meta-operation. We describe this general algorithm next.

\subsection{Converge Algorithm}
The converge algorithm is based on three principles: (1) multiple image classifiers provide multiple views of an image, where each is encoded knowledge in the classifier and can be treated as an instance of a concept; (2) views relate to each other in terms of common abstractions, relationships, or properties that unify them into higher-level knowledge; (3) labels loosely describe a view and so views relate to each other in terms of how their labels relate to each other.

\begin{table}[t]
\vskip -0.15in
\caption{Procedure for unifying views into higher-level knowledge.}
\label{sample-table}
\begin{small}
\begin{center}
\vskip 0.10in
\begin{tabular}{lcccc}
\hline
\abovespace\belowspace
\textbf{Converge (Label: V1, Label: V2, Domain\_Knowledge: DK, Reasoner: R)}\\
	\# initialize to empty set\\
	1: \textbf{Higher\_Level\_Knowledge = \{\}}\\
	\# matches the labels of the classifier with an individual in the ontology\\
	2: \textbf{OWL\_Individual\_1 = Match\_View\_With\_Individual (V1, DK, R)}\\
	3: \textbf{OWL\_Individual\_2 = Match\_View\_With\_Individual (V2, DK, R)}\\
	\# find the common properties between individuals (Properties)\\
	4: \textbf{Common\_Properties =} \\
		\textbf{Get\_Properties (OWL\_Individual\_1, DK, R) $\cap$ Get\_Properties (OWL\_Individual\_2, DK, R)}\\
	\# find binary relationships between these individuals (Relationships)\\
	5: \textbf{Common\_Relationships = Get\_Relationships (OWL\_Individual\_1, OWL\_Individual\_2, DK, R)}\\
	\# find common ancestors (Abstractions)\\
	6: \textbf{Common\_Lowest\_Common\_Ancestor =} 
		\textbf{Lowest\_Level\_Ancestor (}\\
			\textbf{Ancestors (OWL\_Individual\_1, DK, R), Ancestors (OWL\_Individual\_2, DK, R))}\\
	\# add to knowledge\\
    7: \textbf{Higher\_Level\_Knowledge =}\\
	\textbf{Common Properties $\cup$ Common Relationships $\cup$ Common Lowest Common Ancestor}\\
	\# return\\
	8: \textbf{Return Higher\_Level\_Knowledge}\\

 \hline
 
\end{tabular}
\end{center}
\vskip -0.10in
\end{small}

\end{table}

Table 1 gives a high-level walk through of how the converge algorithm unifies views by common properties, abstractions, and relationships. Given two labels \textit{V1} and \textit{V2}, domain knowledge \textit{DK}, and an OWL Reasoner \textit{R}, the algorithm begins at line 1 by instantiating an empty set of knowledge that will ultimately contain the convergences made. Lines 2 and 3 call the \textit{Match\_View\_With\_Individual} function, which maps the views from classifiers onto instances that exist in domain knowledge, \textit{DK} using the views' labels. For example, if \textit{V1} is the string ``banana'' and \textit{V2} is the string ``chimpanzee'', then each call to \textit{Match\_View\_With\_Individual} uses the reasoner \textit{R} and knowledge \textit{DK} to issue a SPARQL \citep{the_sparql_working_group_sparql_2013} query to find an individual named ``banana'' or ``chimpanzee'' and returns the found OWL individual (\textit{OWL\_Individual\_1} and \textit{OWL\_Individual\_2}, respectively).

\begin{table}[t]
\vskip -0.15in
\caption{SMF program output running on three ImageNet pictures.}
\label{sample_input_table}
\begin{small}
\begin{center}
\vskip 0.10in
\begin{tabular}{ >{\centering\arraybackslash}m{1.5in}  >{\centering\arraybackslash}m{.5in} >{\centering\arraybackslash}m{.5in} >{\centering\arraybackslash}m{1in} >{\centering\arraybackslash}m{1in}}
\hline
\abovespace\belowspace
                       Image & ResNet & AlexNet & Explain Output & Converge Output \\
\hline
\abovespace
\vskip -0.15in
\includegraphics[width=1.3in]{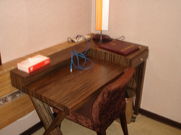}  &   Table lamp  &     Dining table    
&      table lamp is a \textbf{kind of furniture} and dining table is a kind of \textbf{kind of table}   &      \textbf{Furniture} \\
\includegraphics[width=1.3in]{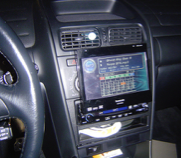} &   CD player  &     Radio    
&      CD player is a \textbf{kind of listening device} and radio is a \textbf{kind of listening device}   &      \textbf{Listening device} \\
\includegraphics[width=1.3in]{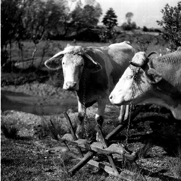}   &   Ox  &     Plow    &      
ox is a \textbf{kind of bovine family} and plow is a \textbf{kind of farming device}   &      Ox \textbf{help farm with} Plow \\

\hline
\end{tabular}
\end{center}
\vskip -0.10in
\end{small}
\end{table}

For both OWL individuals found, Line 4 calls the \textit{Get\_Properties} for \textit{OWL\_Individual\_1} and \textit{OWL\_Individual\_2} and returns their intersection. Each \textit{Get\_Properties} function issues a SPARQL query, using \textit{R} on \textit{DK}, to find any properties that exist in \textit{DK} (either existing in the ontology or inferred by \textit{R}) for an individual and returns the properties. Next, Line 5 calls the \textit{Get\_Relationships} function which finds relationships between both individuals by issuing SPARQL queries for such relationships (either existing in the ontology or inferred by \textit{R}). Continuing with our example, if the relationship \textit{chimpanzee eats banana} is in the ontology, then it will be returned. Next, line 6 finds the lowest common ancestor between two individuals. It does this first by getting the ancestors (either existing in the ontology or inferred by \textit{R}) for each individual by calling the Ancestors function, which, through a recursive call of SPARQL queries (each getting the ancestors of the last ancestors), returns an ordered list of closest to furthest ancestor for the individual. Both lists of ancestors are then passed to the \textit{Lowest\_Level\_Ancestor} function which scans both lists to find the closest ancestor common in both lists. Finally, on lines 7 and 8, the union of properties, relationships, and abstractions is returned. To give an understanding of the knowledge produced from this algorithm, we ran three images from ImageNet through the example program in Figure~\ref{meninx-graph} and provide the output in Table \ref{sample_input_table}, where the \textit{Converge Output} column has the results from running the converge algorithm.

\begin{table}[t]
\vskip -0.15in
\caption{Images mapped to state in the world by classifier and SMF program. We obtained the typewriter from ImageNet and orangutan from Depositphotos (\citeyear{depositphotos_inc_stock_2019}).}
\label{monkey_table}
\begin{small}
\begin{center}
\begin{tabular}{  >{\centering\arraybackslash}m{0.5in}  >{\centering\arraybackslash}m{1.3in} >{\centering\arraybackslash}m{1.3in} >{\centering\arraybackslash}m{1.3in} }
\midrule[1pt]

                       & Initial State P1 & Initial State P2 & Goal State P2 \\
\midrule[1pt]
\vskip -0.15in
Images  &   \includegraphics[width=1.2in]{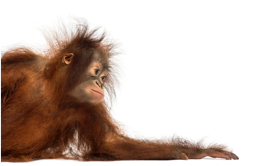}  &    \includegraphics[width=1.2in]{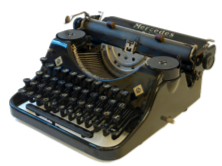}   & \includegraphics[width=1.2in]{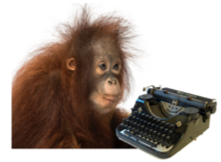} \\
\midrule[1pt]
ResNet Output & orangutan & typewriter & typewriter \\
\midrule[1pt]
AlexNet Output & langur & typewriter & spider monkey \\
\midrule[1pt]
SMF Output & \textbf{primate} & \textbf{typewriter} & spider monkey has typewriter $\Rightarrow$ \textbf{primate has typewriter}\\
\midrule[1pt]
\end{tabular}
\end{center}
\vskip -0.10in
\end{small}
\end{table}

\section{An Example of SMF Program Usefulness}
The ability for a cognitive system, with trained image classifiers, to unify different views of an image into higher-level knowledge should improve the system's ability to formulate and execute plans. In particular, symbolic mirroring finds the level of abstraction on which multiple views agree so the system can avoid asserting incorrect conditions in the world. Establishing correct conditions controls for erroneously executing plans on false preconditions and verifying achievement of postconditions. Combining the SMF program described in Figure \ref{meninx-graph}, with Graphplan \citep{blum_fast_1997}, we demonstrate how symbolic mirroring can take opposing views from classifiers and unify them into a description that correctly describes conditions for planning. This is a crucial ability for a system that has no ground truth to reference, as occurs in test sets.

Suppose, for simplicity, that the system's goal is to instruct a primate to retrieve a nearby typewriter. The first action the system must take is to perceive the world for any primates and typewriters. In this scenario, the system looks into its world at locations P1 and P2 and perceives the images as shown in Table \ref{monkey_table} (Initial State Image P1 and P2). Further, suppose the system has the preconditions and operations as shown in Table \ref{planning_rules_table}'s first two columns for finding a plan using the Graphplan syntax. We now walk through several reasons why symbolic mirroring is important in this scenario.

\begin{table}[t]
\vskip -0.15in
\caption{World facts and planning operators specified in Graphplan.}
\label{planning_rules_table}
\begin{small}
\begin{center}
\vskip 0.10in

\begin{tabular}{ >{\centering\arraybackslash}b{1.8in}|  >{\centering\arraybackslash}b{1.8in}| >{\centering\arraybackslash}b{1.8in}}
\hline
\abovespace\belowspace
                        World Facts & Operators & Plan Found \\
\hline

\begin{flushleft}
(P1)\\
(P2)\\
\medskip
(\textbf{preconds} (\textbf{at} primate P1)\\(\textbf{at} typewriter P2))\\
\medskip
(\textbf{effects} (has-typewriter))\\
\vspace{-3mm}

\end{flushleft}
&
\vspace{-3mm}
\begin{flushleft}
(\textbf{operator}\\
 GOTO\\
 (\textbf{params} (<x>) (<y>))\\
 (\textbf{preconds} (\textbf{at} primate <y>))\\
 (\textbf{effect} (del at primate <y>) \\(\textbf{at} primate <x>)))\\
 \medskip
(\textbf{operator}\\
 GRAB-TYPEWRITER\\
 (\textbf{params} (<y>))\\
 (\textbf{preconds} (\textbf{at} typewriter <y>) (\textbf{at} primate <y>))\\
 (\textbf{effects} (has-typewriter)))\\
 \vspace{-3mm}

\end{flushleft}

&
\begin{flushleft}

\# go to P2 from P1\\
1 GOTO\_P2\_P1\\
\medskip
\# grab typewriter at P2\\
2 GRAB-TYPEWRITER\_P2\\
\vspace{-3mm}

\end{flushleft}
\\

\hline

\end{tabular}

\end{center}
\vskip -0.10in

\end{small}
\end{table}

\textit{SMF Programs Can Generate Plans on General Conditions}. First, suppose our SMF program did not unify different views from classifiers, but rather, suppose it included just one classifier (ResNet or AlexNet) and Graphplan. In the case of ResNet, the program would perceive an orangutan and a typewriter (as indicated in the \textit{ResNet} row in Table \ref{monkey_table}). As such, it could not use a plan that works on primates in general, but rather would need a separate plan for orangutans and for each other type of primate it could possibly predict (at least nine). The same is true for a program that includes only AlexNet, since it is also an ImageNet classifier. We would need to specify many more operators, conditions, and goals for these kinds of programs to generate plans. That is, such programs are unable to generate plans on more general conditions. However, the full SMF program, with symbolic mirroring, will be able to establish the preconditions (as shown in Table \ref{planning_rules_table}) by unifying orangutan and langur into primate.

\textit{SMF Programs Avoid Incorrect Plans}. Second, suppose we added domain knowledge back into the simplified program discussed above. That is, the SMF program is now composed of one classifier (ResNet or AlexNet), Graphplan, and the ImageNet domain knowledge. For the program that is using ResNet, it could use a reasoner to infer that an orangutan is a primate to establish the preconditions and use the operators in Table \ref{planning_rules_table} in order to generate a plan. The same is true for the program that is using AlexNet. However, if the domain knowledge is very detailed and complete (as desired), then many other conditions would also become true. For the ResNet example, plans contingent on the existence of an animal, primate, orangutan, or omnivore all could be tried. For the AlexNet example, plans contingent on the existence of an animal, primate, langur, or herbivore could be attempted. However, this could be disastrous if the classifier does not have an accurate view. For example, a system relying only on AlexNet might execute a feeding plan for langurs (an herbivore) for what is actually an orangutan (an omnivore), which could lead to health problems. However, the full SMF program, with symbolic mirroring, finds the closest abstraction (primate) in knowledge that fits both orangutan and langur, so it would only attempt a plan contingent on primate or an ancestor.

\textit{SMF Programs Can Establish Goal Conditions}. Third, suppose we again have a program composed of one classifier, Graphplan, and domain knowledge. In this scenario, suppose that it produces the plan shown in the \textit{Plan Found} column of Table \ref{planning_rules_table}. Upon executing this plan, the program cannot visually confirm the goal has been obtained. After executing line 2 of the plan (GRAB-TYPEWRITER\_P2), the program using ResNet can only confirm that a typewriter is at P2 as shown in Table \ref{monkey_table}. The program using AlexNet will assert that a spider monkey is at P2 as shown in Table \ref{monkey_table}. However, the SMF program can unify both classifier's views into the assertion that a spider monkey has a typewriter through the relationship \textit{has} in its domain knowledge. The system could then use this relationship to infer that a primate has a typewriter, because spider monkey has the ancestor class primate.

\section{Exploratory Studies}
As a first step in understanding the process of developing systems in our SMF, how they perform on large data sets, and the generality of symbolic mirroring in them, we ran an exploratory study on three different programs. They are all variations on Figure \ref{meninx-graph}, where program RA uses classifiers ResNet and AlexNet, RS uses ResNet and SqueezeNet, and AS uses AlexNet and SqueezeNet. While there are many classifiers to choose from, we selected ResNet, AlexNet, and SqueezeNet to see if larger or smaller differences in the top--1 accuracies (77.11, 54.92, 56.11) had an effect on the number of different views. Further, they are some of the most well-known and evaluated image classifiers in research. We provided each program with the meta-knowledge and domain knowledge described in Figure \ref{planning_rules_table}. We also performed a brief field study, where we fed live images to an SMF program while walking around. This tested the system on a different data distribution and provided an idea of the practicality of using SMF in real environments.

\subsection{Study Designs}
In order to mimic some of the situations in which a SMF system might be created, we created RA, RS, and AS under five design conditions:
\vskip 0.05in

\cbullet 
The author who created the domain knowledge (as shown in Figure 4) was given the thousand ImageNet labels beforehand and approximately three days to create domain knowledge in OWL using Prot\'eg\'e. This placed a realistic constraint on how complete the knowledge could be before the program was run. 
\cbullet 
Another author chose 950 pictures distributed by ImageNet into the categories of animals (250), electronics (250), food (250), and furniture (200). 
\cbullet 
The categories were shown to the author of the domain knowledge (but not the pictures) so that he might focus some of his attention here, but there was no strict requirement. The author of the knowledge also received 20 example images for each category to help in creating relationships, properties, or abstractions. 
\cbullet 
While the abstractions and properties were based on possible labels from a classifier, the relationships were inferred by the author who created the domain knowledge by looking at example images. These relationships might not actually exist in the test set, a concern that we address with an evaluation in Section 4.4.
\cbullet 
The classifiers used in these programs were ``off the shelf'' from Gluon \citep{the_apache_software_foundation_gluon_2019} and were not specially trained for this experiment. This helped us evaluate symbolic mirroring orthogonal to learning.
\vskip 0.05in
\noindent
For the field study, we extended RA in Figure 3 so that the images fed through it were from a live video camera and we added a speech synthesizer so that it would speak out loud about what views it unified. We did this by adding a few more nodes to the program graph of RA. We walked around in an office setting and recorded the number of unified views.

\begin{table}[t]
\vskip -0.15in
\caption{Explained and unified results (percentages rounded) in laboratory study.}
\label{lab_results_table}
\begin{small}
\begin{center}
\vskip 0.10in
\setlength\tabcolsep{4pt} 

\begin{tabular}{ >{\centering\arraybackslash}m{.2in} >{\centering\arraybackslash}m{.4in}  >{\centering\arraybackslash}m{.6in} >{\centering\arraybackslash}m{.6in} >{\centering\arraybackslash}m{.6in}
>{\centering\arraybackslash}m{.6in} >{\centering\arraybackslash}m{.6in} >{\centering\arraybackslash}m{.6in}
>{\centering\arraybackslash}m{.6in}}

\toprule[1pt]

ID & Top--1 Acc. Diff. & Different Views Total & Unified Total & Disunited Total & Explained
Total & Not Explained Total & Unified Total in Explained & Disunited Total in Explained \\
\midrule[1pt]

RA & 22.2 & 369
(39\%) &	159
(43\%)&	210
(57\%)&	239
(65\%)&	130
(35\%)&	159
(67\%)&	80
(33\%) \\
\midrule[1pt]

RS & 21 & 384
(40\%) & 158
(41\%) & 226
(59\%) & 220
(57\%) & 164
(43\%) & 158
(72\%) & 62
(28\%) \\
\midrule[1pt]
AS & 1.2 & 349
(37\%) & 161
(46\%) & 188
(54\%) & 215
(62\%) & 134
(38\%) & 161
(75\%) & 54
(25\%)\\
\midrule[1pt]

\end{tabular}
\end{center}
\vskip -0.10in
\end{small}
\end{table}

\subsection{Study Results}
We next present the results from the laboratory and field studies. We measured the performance of the programs by the percentage of unified views and explanations they created. We also include results from a survey conducted to measure if the unified views made by a SMF program actually describe something in the images fed to it (a type of ``accuracy''). We include significant ($\alpha = .05$) $p$ values by applying Chi-squared on 2$\times$1 contingency tables.

Table \ref{lab_results_table} presents the results for programs RA, RS, and AS. The \textit{Top--1 Acc. Diff.} column provides the differences between the reported top--1 accuracies for the classifiers used in the programs. \textit{Different Views Total} provides the total amount of views that differ by label created from 950 different images, \textit{Unified Total} gives the number of different views unified, and \textit{Disunited Total} reports the total number of differing views that were not unified. \textit{Explained Total} gives the number of explained differences between views, while \textit{Unexplained Total} provides the number of views that could not be explained because they had no mapping into knowledge (i.e., there are missing classes in the knowledge base). \textit{Unified Total in Explained} gives the number of views unified among the explained and \textit{Disunited Total in Explained} gives the number of views not unified among the explained. We do not present
\textit{Unified Total in Unexplained} because they were 0\% for each row, which means that, if differing views were unexplained, they could not be unified.

We found that between 41\% to 46\% of views were unified and between 54\% to 59\% remained disunited (p $\leq$ .001 for RS and p $\leq$ .008 for RA). We also found that 57\% to 65\% of different views were explained for each program (all p $\leq$ .004), and that among these 67\% to 75\% were unified for each program (all p $\leq$.001). In addition, we found that the accuracy differences between the classifiers used in each program have only a small impact on the number of different views produced (the maximum difference between the quantity of different views was 3\%).

\begin{table}[t]
\vskip -0.15in
\caption{Frequency of unified views by type results (percentages rounded) from laboratory study.}
\label{lab_results_type_table}
\begin{small}
\begin{center}
\vskip 0.10in

\begin{tabular}{ >{\centering\arraybackslash}m{.2in} >{\centering\arraybackslash}m{.8in}  >{\centering\arraybackslash}m{.8in} >{\centering\arraybackslash}m{.8in} >{\centering\arraybackslash}m{.8in}
>{\centering\arraybackslash}m{.8in} >{\centering\arraybackslash}m{.8in}}
\toprule[1pt]

ID & Abstractions Total &  Properties Total & Relationships Total & Multiple Unified Total & Unified Total \\
\midrule[1pt]

RA & 149 (79\%) & 30 (16\%) & 10 (5\%) & 30 (16\%) & 189 \\ 
\midrule[1pt]
RS & 143 (78\%) & 26 (14\%) & 15 (8\%) & 25 (14\%) & 184 \\
\midrule[1pt]
AS & 149 (78\%) & 31 (16\%) & 12 (6\%) & 33 (17\%) & 192 \\
\midrule[1pt]

\end{tabular}
\end{center}
\vskip -0.10in
\end{small}
\end{table}

Table \ref{lab_results_type_table} gives the breakdown among the unified views (159, 158, 161) in terms of the kinds of convergences that happened by abstraction, properties, and relationships. In some cases, views were unified by more than one method, as given in the \textit{Multiple Unified Total} column: this is why the \textit{Unified Total} column has numbers greater than 159, 158, and 161, respectively. We found the most common way to unify views was through abstraction (i.e., lowest common ancestors), followed by unifying through common properties and relationships. We present a few results from each category in  Table \ref{sampe_results_table} from RA as illustrations.

\begin{figure}[b]

\vskip 0.05in
\begin{center}
\includegraphics[width=5.5in]{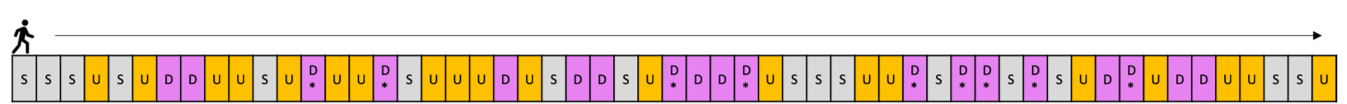}
\caption{Ordered encoding of unified, disunited, and explained views in field study results.} 
\label{field_study}
\end{center}
\end{figure}

For the field study, a person held a camera that fed images from an office environment into the RA program (as described in Figure \ref{meninx-graph}) but modified so that it took live video images. In total, the system processed 55 images from the camera. Figure \ref{field_study} presents the ordering of these results, where ``S'' means no differing views were generated, ``U'' means differing views were unified, ``D'' means differing views were not unified, and ``D*'' means differing views that had an explanation were not unified. We found 39 (71\%) differing views occurred. Of the 39 differences, 20 were unified (51\%) and 19 (49\%) were not. Some 29 (74\%) of the differing views had explanations (counting ``U'' and ``D*''). Of these, 20 views (69\%) were unified and nine (31\%) remained disunited ($p = .041$).

\begin{table}[t]
\vskip -0.15in
\caption{Three ImageNet examples from laboratory study by each type of convergence.}
\label{sampe_results_table}
\begin{small}
\begin{center}
\vskip 0.10in
\begin{tabular}{ >{\centering\arraybackslash}m{1.5in}  >{\centering\arraybackslash}m{.5in} >{\centering\arraybackslash}m{.5in} >{\centering\arraybackslash}m{1in} >{\centering\arraybackslash}m{1in}}
\hline
\abovespace\belowspace
                       Image & ResNet & AlexNet & Explain Output & Converge Output \\
\hline
\abovespace
\vskip -0.15in
\includegraphics[width=1.4in]{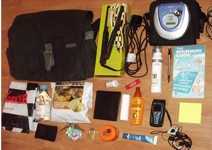}  &   Backpack  &     Purse    
&      backpack is a \textbf{kind of carrying device} and purse \textbf{is a kind of carrying device}   &      \textbf{Carrying device}
(abstraction)
 \\
\includegraphics[width=1.4in]{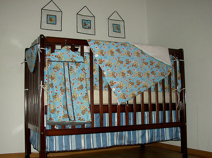} &  Crib  & Quilt   
&      crib is a \textbf{kind of bed} and quilt is a \textbf{kind of bedding}   &      Quilt \textbf{lays on furniture} crib
(relationship)
 \\
\includegraphics[width=1.4in]{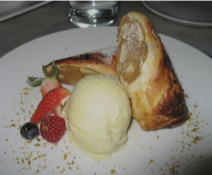}   &   Ice Cream  &     Mashed Potato    &      
ice cream is a \textbf{kind of sweets} and mashed potato is a \textbf{kind of prepared potatoes}   &      \textbf{Is edible}
(property)

\textbf{Food family}
(abstraction)
 \\
\hline
\end{tabular}
\end{center}
\vskip -0.10in
\end{small}
\end{table}

\subsection{Survey Results on Unified Views}
In order to understand if the unified views actually make sense as descriptions of images, we conducted a survey to evaluate a kind of ``accuracy'' for the program RA. Traditionally, accuracy of a classifier is measured before deployment with a test set, but an SMF program uses classifiers that have already been deployed and the system is being used regularly.
\nowidow
In an online survey, we asked 24 people (three authors and 21 volunteers recruited over a company channel) to evaluate if unified views described their corresponding images or not. Specifically, we asked each participant \textit{Does the word or phrase listed to the left of each picture describe something in the picture?} for 40 unified views and their corresponding images. They indicated their answer with a button labeled ``agree'' or ``disagree.'' For each participant, 10 unified views made by program RA, were chosen at random from each category (furniture, food, electronics, and animals), so all categories were represented. We decided to ask only 40 questions, rather than 159 for all unified views made by RA, because the length of the survey would then have been prohibitively long. In total, we asked 960 questions across the 24 participants.

We found that for 839/960 (87\%) of the questions, the participants agreed the unified views described something in the image ($p \leq .001$). By category, participants agreed the unified views described their corresponding images for 222/240 (92.5\%) of the animal images ($p \leq .001$), 209/240 (87\%) of the furniture images ($p \leq .001$), 203/240 (84.5\%) of the food images ($p \leq .001$), and 205/240 (85.4\%) of the electronic images ($p \leq .001$).

\subsection{Threats to Study Validity}
While our results are promising, there are threats to its validity.  For external validity, our study is limited to a common-sense domain (ImageNet labels), where we did not seek expert opinion on the domain knowledge created. Thus, the results may not apply to expert areas, but common-sense reasoning in cognitive systems is important \citep{davis_commonsense_2015}. Further, while the laboratory study provided control for us to take measurements without noise, we had to give up the realism of the study by using the ImageNet data set. However, we attempted to compensate for this with a field study that demonstrated the approach in a real environment on a different data distribution. Since the laboratory and field study told similar stories, we are confident in our results.

For internal validity, the abstractions and properties in the domain knowledge were inferred from the labels of the classifiers that were known to exist in the test set. However, by the developer inferring the relationships from the few example images he had might mean they do not exist in the test set. To see if they did, we conducted a survey among seven people (four authors and three volunteer, five male and two female) who took an online survey presenting each with a relationship produced from the laboratory study, the associated image, and a choice of ``relationship exists'' or ``relationship does not exist'' (creating 37 questions total). We asked them to use their own judgement when choosing and we provided no definition of relationship. We found that, for 29 (78\%) of the images, most participants agreed the relationship existed in the image. This gives us confidence that, when a relationship was used to unify views, the relationship likely existed.

\section{Discussion}
The results suggest our approach and framework can help build cognitive systems capable of unifying views into higher-level knowledge after learning. While 54\% to 59\% of results remained disunited in the laboratory study, the results also showed that 41\% to 46\% were unified in the laboratory study and 51\% unified in the field study. For systems processing thousands to millions of images, this could have a large impact. Further, the system greatly improves when differing views were explained (i.e., views could be mapped into the domain knowledge classes), where the system could unify 67\% to 75\% of views in the laboratory study and 69\% in the field study. This suggests that a cognitive system built with our SMF can gain a unified understanding of 41\% to 75\% of images being processed (possibly millions). 

The survey results (87\% of unified views are descriptive) suggest that the symbolic mirroring approach often produced accurate descriptions. However, it also means that views with incorrect labels were unified into words or phrases people thought were descriptive of the corresponding images. This implies that the symbolic mirroring approach is a means to describe the knowledge encoded in classifiers themselves and could help reveal what is learned in a classifier rather than relying only on the classifier's predicted label.

The results also suggest two important and potentially powerful developer methods of improving cognitive systems. First, as we saw in Table \ref{lab_results_table}, if the system could explain differing views, then its chance of unifying them improved greatly. This suggests that developers could continually add instances of concepts to the domain knowledge every time their program encountered a view from a classifier that it had not encountered before. Tools for recording when the cognitive system does not know how to unify views or is ``confused'' will be needed. Second, as discussed in Subsection 4.4, the author of the domain knowledge could infer relationships from relatively few examples and these relationships often existed in the test set. This suggests developers could add relationships to domain knowledge with relatively few image examples, an interesting contrast to the millions of examples used by machine learning today.

In the laboratory study setting, the author creating the domain knowledge was given the four categories animals, electronics, food, and furniture to help focus his development. However, many of the views in the experiment did not map into these categories, as seen by the number not explained because of missing knowledge. This suggests that differing views from classifiers can vary widely across categories, which means that a ``conceptually wide'' domain knowledge is needed to account for conceptual differences among the different views.

\section{Related Work}
Our research contrasts with previous research in meta-reasoning along two dimensions. First, we investigate meta-reasoning as a sense-making mechanism to unify the visual world by bridging neural networks with symbolic approaches. Second, we provide a modular framework for using meta-reasoning and meta-operations in a way more flexible than the classic approach.

Meta-reasoning has been described as an essential component of general intelligence, where it is often argued that it is critical for choosing different ways to think given finite resources, as in humans \citep{epstein_learning_2011, griffiths_doing_2019}. One example is changing game strategies in new environments. For example, adaptation in response to the external environment has been demonstrated by agents reasoning over models of their own strategies \citep{goel_metareasoning_2011, rugaber_gaia_2013} and altering those models to meet external demands. One aspect of these agents is their representation of themselves in the Task Method Knowledge Language \citep{murdock_meta-case-based_2008}, which lets the agent reason over its goals and methods. Similarly, meta-reasoning has been applied to self-monitoring and self-repair when anomalous situations arise \citep{schmill_metacognitive_2011}. From an interactive perspective, a question and answer system has been designed to answer ``common sense'' kinds of questions about itself \citep{morbini_metareasoning_2011}. This approach is particularly appealing form a debugging point of view, where explanations can provide understanding and link events \citep{cox_metareasoning_2011}. In image processing, work on pipeline selection has used meta-reasoning for choosing alternatives based on context \citep{robertson_metareasoning_2011}, much as in Auto ML \citep{feurer_efficient_2015}. In contrast, our work applies meta-reasoning to unify viewpoints of the visual world into higher-level knowledge structures that the system can use in a variety of applications (e.g., planning or explanation). In particular, in Section 3, we demonstrated how our approach can be used to avoid incorrect conditions in planning and detect goals. Further, it is a step closer to building systems similar to how humans think in a more unified way \citep{bayne_what_2003, shivhare_cognitive_2016} than the disjoint predictions of image classifiers.

\cite{gilpin_monitoring_2018} demonstrated a monitoring technique to analyze captions from image classifiers and determine if they are reasonable. The method involved mapping each part of speech from the captions onto slots in frames that represent primitive actions. In contrast, our work is at the lower level of the perception. In particular, symbolic mirroring unifies different views in terms of abstractions, properties, or relationships, which can then be monitored for reasonableness with systems like those demonstrated by \citeauthor{gilpin_monitoring_2018}

While there are a variety applications for meta-reasoning, architectures are often presented in a uniform way that includes a ground level, an object level, and a meta-level \citep{russell_principles_1991-1}. Yet, this separation has an impact in the design of these systems in two ways. First, operations in each level must be treated as one type, but common reasoning methods will often cross cut layers. Indeed, some have claimed ``\dots metareasoning and reasoning are entangled in such a way that it is impossible to separate'' \citep{robertson_metareasoning_2011}. Second, the runtime system must have three separate modes and one intermediate mode -- making it hard to seamlessly switch between levels. This also has the consequence that the execution of the ground or object levels are separate (making it harder to intervene and control execution from the meta-level). In this work, we introduced meta-points and meta-operations as nodes in a graph, where each node can be any arbitrary function that switch from any level at any execution point.

\section{Conclusion}
In this paper, we identified the problem of unifying different views of classifiers into high-level knowledge after learning. We developed a general approach, symbolic mirroring, and a framework that leverages meta-reasoning in order to unify different views from classifiers into higher-level symbolic knowledge. Through a planning example, we demonstrated how an SMF program avoids generating plans based on incorrect assertions, how it supports general planning operations, and how it helps identify goal conditions.  In an exploratory laboratory study, we found that SMF programs unified 41\% to 46\% of the views of images and, among the views that could be explained, unified 67\% to 75\% of these. Further, in an exploratory field study, we found the SMF program unified 51\% of the views from live video images and, among those that could be explained, the system could unify 69\% of them, showing the approach can work on a different data distribution than ImageNet and ``in the wild''. Further, a survey of 24 people suggested that 87\% of the unified views made by the SMF program described the images it processed.

Our future work will proceed in two directions. From the lower level, we will investigate what kinds of predictions from image classifiers work better in our approach. Rather than using image classifiers that predicate complex objects (e.g., dog or jacket), we will see if simpler predictions (e.g., edge or square) from different classifiers can be unified into more complex objects using symbolic mirroring. In a sense, we will be investigating the balance between the complexity of predictions with the complexity of knowledge and reasoning needed to unify predictions. From the higher level, we will research how to unify abstractions, relationships, and properties into higher-level stories that have temporal components (as in the field study) or that are created from a single image. In particular, we will see how to take unified views from sequential video frames and unify them into descriptive sentences. Applications could extend to letting a system understand the ethical implications of a story it creates from images.

\begin{acknowledgements} 
\noindent
We would like to thank the MIT-IBM Watson AI Lab, Nicola Palmarini, Grady Booch, and the reviewers for their support and insight in shaping this research.
\end{acknowledgements} 

\vspace{-0.25in}

{\parindent -10pt\leftskip 10pt\noindent
\bibliographystyle{cogsysapa}
\bibliography{references_cogsys.bib}

}


\end{document}